\documentclass[11pt,a4paper]{article}
\usepackage[hyperref]{emnlp2018}
\usepackage{times}
\usepackage{latexsym}

\usepackage{url}
\usepackage{graphicx,amsmath,amssymb,threeparttable,url,subfig,placeins} 

\usepackage{listings}
\usepackage{xcolor}
\usepackage{varwidth}
\usepackage{cprotect}

\aclfinalcopy 

\title{CytonMT: an Efficient Neural Machine Translation\\
Open-source Toolkit Implemented in C++
}

\author{Xiaolin Wang \hspace{.2cm} Masao Utiyama \hspace{.2cm}  Eiichiro Sumita\\
Advanced Translation Research and Development Promotion Center \\
National Institute of Information and Communications Technology, Japan\\
{\tt \{xiaolin.wang,mutiyama,eiichiro.sumita\}@nict.go.jp}
}

\date{}

\begin{document}
\maketitle
\begin{abstract}
This paper presents an open-source neural machine translation toolkit named CytonMT\footnote{https://github.com/arthurxlw/cytonMt}. The toolkit is built from scratch only using C++ and NVIDIA's GPU-accelerated libraries. The toolkit features training efficiency, code simplicity and translation quality. Benchmarks show that CytonMT accelerates the training speed by 64.5\% to 110.8\% on neural networks of various sizes, and achieves competitive translation quality.
\end{abstract}

\section{Introduction}

Neural Machine Translation (NMT) has made remarkable progress over the past few years~\cite{sutskever2014sequence,bahdanau2014neural,wu2016google}.  Just like Moses~\cite{koehn2007moses} does for statistic machine translation (SMT), open-source NMT toolkits contribute greatly to this progress, including but not limited to,
\begin{itemize}
\item RNNsearch-LV~\cite{jean2015using}\footnote{https://github.com/sebastien-j/LV\_groundhog}
\item Luong-NMT~\cite{luong2015effective}\footnote{https://github.com/lmthang/nmt.hybrid}
\item DL4MT by Kyunghyun Cho et al.\footnote{https://github.com/nyu-dl/dl4mt-tutorial}
\item BPE-char~\cite{chung2016character}\footnote{https://github.com/nyu-dl/dl4mt-cdec}
\item Nematus~\cite{sennrich2017nematus}\footnote{https://github.com/EdinburghNLP/nematus}
\item OpenNMT~\cite{klein2017opennmt}\footnote{https://github.com/OpenNMT/OpenNMT-py}
\item Seq2seq~\cite{britz2017massive}\footnote{https://github.com/google/seq2seq}
\item ByteNet~\cite{DBLP:journals/corr/KalchbrennerESO16}\footnote{https://github.com/paarthneekhara/byteNet-tensorflow (unofficial) and others.}
\item ConvS2S~\cite{gehring2017convs2s}\footnote{https://github.com/facebookresearch/fairseq}
\item Tensor2Tensor~\cite{vaswani2017attention}\footnote{https://github.com/tensorflow/tensor2tensor}
\item Marian~\cite{junczys2018marian}\footnote{https://github.com/marian-nmt/marian}
\end{itemize}

These open-source NMT toolkits are undoubtedly excellent software.  However, there is a common issue -- they are all written in script languages with dependencies on third-party GPU platforms (see Table~\ref{tab:toolkit}) except Marian which is developed simultaneously with our toolkit.

Using script languages and third-party GPU platforms is a two-edged sword. On one hand, it greatly reduces the workload of coding neural networks. On the other hand, it also causes two problems as follows,

\begin{itemize}
\item The running efficiency drops, and profiling and optimization also become difficult, as the direct access to GPUs is blocked by the language interpreters or the platforms. NMT systems typically require days or weeks to train, so training efficiency is a paramount concern. Slightly faster training can make the difference between plausible and impossible experiments~\cite{klein2017opennmt}.

\item The researchers using these toolkits may be constrained by the platforms. Unexplored computations or operations may become disallowed or unnecessarily inefficient on a third-party platform, which lowers the chances of developing novel neural network techniques. 
\end{itemize}

\begin{table}
{
\small
\begin{center}
\begin{tabular}{|l|ll|r|}
\hline
Toolkit & Language & Platform\\
\hline
RNNsearch-LV & Python & Theano,GroundHog \\
Luong-NMT & Matlab& Matlab \\
DL4MT & Python & Theano \\
BPE-char & Python & Theano \\
Nematus   & Python & Theano \\
OpenNMT    & Lua & Torch \\
Seq2seq    & Python & Tensorflow \\
ByteNet & Python & Tensorflow \\
ConvS2S & Lua & Torch \\
Tensor2Tensor & Python &Tensorflow \\
\hline
Marian           & C++ & -- \\
CytonMT          & C++ & -- \\
\hline
\end{tabular}
\end{center}
}
\caption{\label{tab:toolkit} Languages and Platforms  of Open-source NMT toolkits 
}
\end{table}

CytonMT is developed to address this issue, in hopes of providing the community an attractive alternative. The toolkit is written in C++ which is the genuine official language of NVIDIA -- the manufacturer of the most widely-used GPU hardware. This gives the toolkit an advantage on efficiency when compared with other toolkits. 


Implementing in C++ also gives CytonMT great flexibility and freedom on coding. The researchers who are interested in the real calculations inside neural networks can trace source codes down to kernel functions, matrix operations or NVIDIA's APIs, and then modify them freely to test their novel ideas.
  
The code simplicity of CytonMT is comparable to those NMT toolkits implemented in script languages. This owes to an open-source general-purpose neural network library in C++, named CytonLib, which is shipped as part of the source code. The library defines a simple and friendly pattern for users to build arbitrary network architectures in the cost of two lines of genuine C++ code per layer.

CytonMT achieves competitive translation quality, which is the main purpose of NMT toolkits. It implements the popular framework of attention-based RNN encoder-decoder. Among the reported systems of the same architecture, it ranks at top positions on the benchmarks of both WMT14 and WMT17 English-to-German tasks.

The following of this paper presented the details of CytonMT from the aspects of method, implementation, benchmark, and future works.

\section{Method}

\begin{figure}[tb!]
\begin{center}
\includegraphics[width=.49\textwidth]{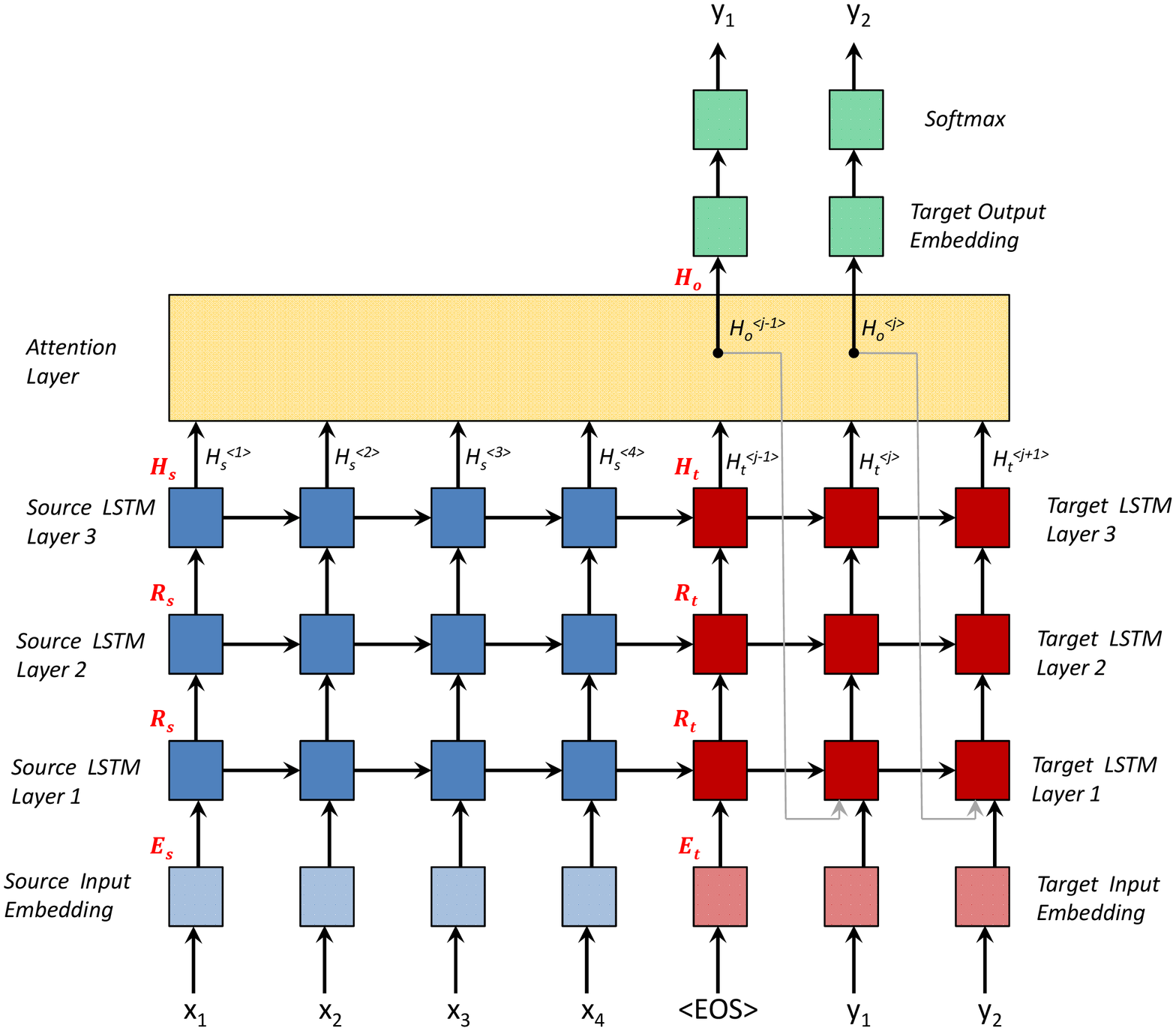} 
\end{center}
\caption{Model Architecture of CytonMT}
\label{fig:architecture}
\end{figure}

The toolkit approaches to the problem of machine translation using the attention-based RNN encoder-decoder proposed by \citet{bahdanau2014neural} and \citet{luong2015effective}. The figure~\ref{fig:architecture} illustrates the architecture. The conditional probability of a translation given a source sentence is formulated as,
{\small
\begin{alignat}{2}
log\mathop{p} ( \mathbf{y}|\mathbf{x} ) 
  & = \sum_{j=1}^m log( \mathop{p}(y_j|H_o^{ \langle j \rangle }) \nonumber \\
    = \sum_{j=1}^m \mathrm{log} & (\mathrm{softmax}_{y_j}( \mathrm{tanh} ( W_o H_o^{ \langle j \rangle }+B_o))) \\
H_o^{ \langle j \rangle } & = \mathcal{F}_\mathrm{att}(H_s, H_t^{ \langle j \rangle }) , 
\end{alignat}
}
where $\mathbf{x}$ is a source sentence;
$\mathbf{y}$=$(y_1, \ldots, y_m)$ is a translation; 
$H_s$ is a source-side top-layer hidden state;
$H_t^{\langle j \rangle }$ is a target-side top-layer hidden state;
$H_o^{\langle j \rangle }$ is a state generated by an attention model $ \mathcal{F}_\mathrm{att}$; $W_o$ and $B_o$ are the weight and bias of an output embedding.

\begin{figure}[tb!]
\begin{center}
\vspace{25pt}
\includegraphics[width=.49\textwidth]{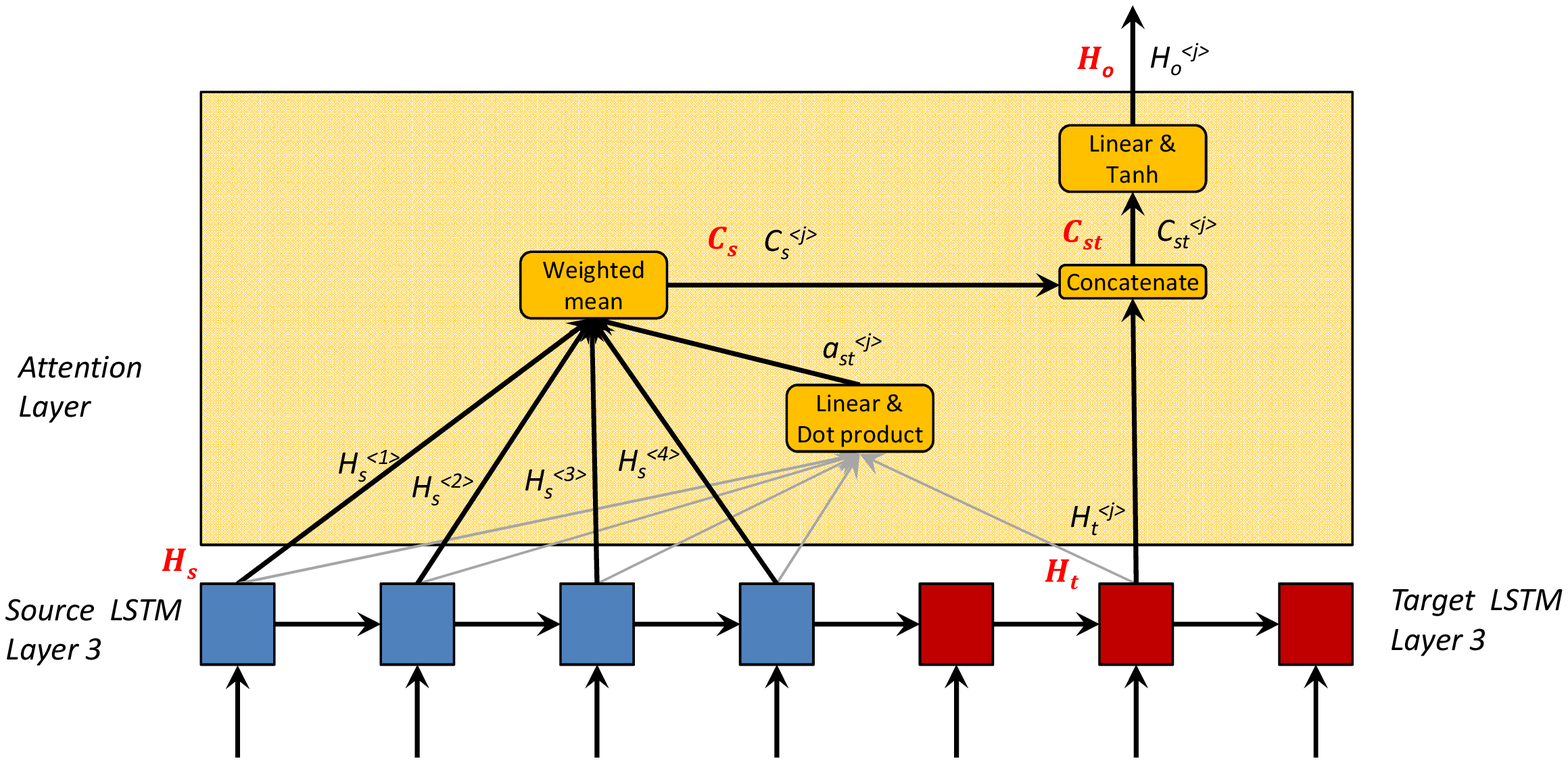}
\end{center}
\caption{Architecure of Attention Model}
\label{fig:attention}
\end{figure}

The toolkit adopts the multiplicative attention model proposed by \citet{luong2015effective}, because it is slightly more efficient than the additive variant proposed by \citet{bahdanau2014neural}. This issue is addressed in \citet{britz2017massive} and \citet{vaswani2017attention}. The figure~\ref{fig:attention} illustrates the model, formulated as ,
{\small
\begin{alignat}{2}
a_{st}^{ \langle i j \rangle } & =  \mathrm{softmax} (\mathcal{F}_\mathrm{a}(H_s^{\langle i \rangle }, H_t^{\langle j \rangle })) \nonumber   \\
      & =    
\frac{
		\mathop{e}^ {\mathcal{F}_\mathrm{a}(H_s^{\langle i \rangle }, H_t^{\langle j \rangle })}}
		{\sum_{i=1}^n \mathop{e} ^ {\mathcal{F}_\mathrm{a}(H_s^{\langle i \rangle }, H_t^{\langle j \rangle })} 
} \label{eq:a_st} , \\
\mathcal{F}_\mathrm{a}(H_s^{\langle i \rangle }, H_t^{\langle j \rangle }) & =  H_s^{\langle i \rangle }{}^{\top} W_\mathrm{a} H_t^{\langle j \rangle } ,\label{f_att} \\
C_s^{\langle j \rangle } & =  \sum_{i=1}^n a_{st}^{\langle i j \rangle}  H_s^{\langle i \rangle}, \label{eq:c_s} \\
C_{st}^{\langle j \rangle } & =  [ C_s^{} ; H_t^{\langle j \rangle } ] , \label{eq:c_st}  \\
H_o^{\langle j \rangle } & =  \mathrm{tanh} ( W_c C_{st}^{\langle j \rangle }), \label{eq:h_o} 
\end{alignat}
}
where $\mathcal{F}_\mathrm{a}$ is a scoring function for alignment; $W_\mathrm{a}$ is a matrix for linearly mapping target-side hidden states into a space comparable to the source-side; $a_{st}^{\langle i j \rangle}$ is an alignment coefficient; $C_s ^{\langle j \rangle }$ is a source-side context; $C_{st} ^{\langle j \rangle }$ is a context derived from both sides.

\section{Implementation}

The toolkit consists of a general purpose neural network library, and a neural machine translation system built upon the library. The neural network library defines a class named {\it Network} to facilitate the construction of arbitrary neural networks. Users only need to inherit the class, declare components as data members, and write down two lines of codes per component in an initialization function.  For example, the complete code of the attention network formulated by the equations \ref{eq:a_st} to \ref{eq:h_o} is presented in the figure~\ref{fig:code:attention}. This piece of code fulfills the task of building a neural network as follows,
\begin{itemize}

\item The class of {\it Variable} stores numeric values and gradients. Through passing the pointers of {\it Variable} around, component are connected together.

\item The data member of {\it layers} collects all the components. The base class of {\it Network} will call the functions {\it forward}, {\it backward} and {\it calculateGradient} of each component to perform the actual computation.

\end{itemize}

\begin{figure}
\begin{center}
\noindent\cprotect\fbox{\begin{minipage}{0.45\textwidth}
{\scriptsize
\begin{verbatim} 
class Attention: public Network
{
 DuplicateLayer dupHt;      // declare components
 LinearLayer linearHt;
 MultiplyHsHt multiplyHsHt;
 SoftmaxLayer softmax;
 WeightedHs weightedHs;
 Concatenate concateCsHt;
 LinearLayer linearCst;
 ActivationLayer actCst;

 Variable* init(LinearLayer* linHt, 
    LinearLayer* linCst, Variable* hs,
    Variable* ht)
 {
  Variable* tx;
  tx=dupHt.init(ht);           // make two copies
  layers.push_back(&dupHt);

  tx=linearHt.init(linHt, tx);            // WaHt
  layers.push_back(&linearHt);

  tx=multiplyHsHt.init(hs, tx);             // Fa
  layers.push_back(&multiplyHsHt);

  tx=softmax.init(tx);                     // ast
  layers.push_back(&softmax);

  tx=weightedHs.init(hs, tx);               // Cs
  layers.push_back(&weightedHs);

  tx=concateCsHt.init(tx, &dupHt.y1);      // Cst
  layers.push_back(&concateCsHt);

  tx=linearCst.init(linCst, tx);         // WcCst
  layers.push_back(&linearCst);

  tx=actCst.init(tx, CUDNN_ACTIVATION_TANH);// Ho
  layers.push_back(&actCst);

  return tx; //pointer to result
 }
};
\end{verbatim}
}
\end{minipage}}
\end{center}
\caption{Complete Code of Attention Model Formulated by Equations \ref{eq:a_st} to \ref{eq:h_o} }
\label{fig:code:attention}
\end{figure} 

The codes of actual computation are organized in the functions {\it forward}, {\it backward} and {\it calculateGradient} for each type of component. The figure~\ref{fig:code:working} presents some examples. Note that these codes have been slightly simplified for illustration.

\begin{figure}
\begin{center}
\noindent\cprotect\fbox{\begin{minipage}{0.45\textwidth}
{\scriptsize
\begin{verbatim}

void LinearLayer::forward()
{
 cublasXgemm(cublasH, CUBLAS_OP_T, CUBLAS_OP_N,
  dimOutput, num, dimInput,
  &one, w.data, w.ni, x.data, dimInput,
  &zero, y.data, dimOutput)
}

void LinearLayer::backward()
{
 cublasXgemm(cublasH, CUBLAS_OP_N, CUBLAS_OP_N,
  dimInput, num, dimOutput,
  &one, w.data, w.ni, y.grad.data, dimOutput,
  &beta, x.grad.data, dimInput));
}

void LinearLayer::calculateGradient()
{
 cublasXgemm(cublasH, CUBLAS_OP_N, CUBLAS_OP_T,
  dimInput, dimOutput,  num,
  &one, x.data, dimInput, y.grad.data, dimOutput,
  &one, w.grad.data, w.grad.ni));
}

void EmbeddingLayer::forward()
{
 ...
 embedding_kernel<<<grid, blockSize>>>(words, 
  firstOccurs, len, dim, stride,
  wholeData, y.data, true);
}

\end{verbatim}
}
\end{minipage}}

\end{center}
\caption{Codes of Performing Actual Computation.  }
\label{fig:code:working}
\end{figure} 

\section{Benchmarks}
\label{sec:experiments}

\subsection{Settings}
CytonMT is tested on the widely-used benchmarks of the WMT14 and WMT17 English-to-German tasks~\cite{bojar-EtAl:2017:WMT1} (Table~\ref{tab:dataset}). Both datasets are processed and converted using byte-pair encoding\citep{gage1994new,schuster2012japanese} with a shared source-target vocabulary of about 37000 tokens.  The WMT14 corpora are processed by the scripts from \citet{vaswani2017attention}\footnote{https://github.com/tensorflow/tensor2tensor}. The WMT17 corpora are processed by the scripts from  \citet{junczys2018marian}\footnote{https://github.com/marian-nmt/marian-examples/tree/master/wmt2017-uedin}, which includes 10 million back-translated sentence pairs for training.


The benchmarks were run on an Intel Xeon CPU E5-2630 @ 2.4Ghz and a GPU Quadro M4000 (Maxwell) that had 1664 CUDA cores @ 773 MHz, 2,573 GFLOPS . The software is CentOS 6.8, CUDA 9.1 (driver 387.26), CUDNN 7.0.5, Theano 1.0.1, Tensorflow 1.5.0. Netmaus, Torch and OpenNMT are the latest version in December 2017. Marian is the last version in May 2018.


CytonMT is run with the hyperparameters settings presented by Table~\ref{tab:hparam} unless stated otherwise. The settings provide both fast training and competitive translate quality according to our experiments on a variety of translation tasks. Dropout is applied to the hidden states between non-top recurrent layers $R_s$, $R_t$ and output $H_o$ according to ~\citep{wang2017empirical}. Label smoothing estimates the marginalized effect of label-dropout
during training, which makes models learn to be more unsure~\cite{szedegy2016rethinking}. This improved BLEU scores~\cite{vaswani2017attention}. Length penalty is applied using the formula in~\cite{wu2016google}.

\begin{table}
{\small
\begin{center}
\begin{tabular}{|lrrr|}
\hline
\bf Data Set & \bf \# Sent. & \multicolumn{2}{c|}{\bf \# Words}   \\ 
             &                  & Source & Target     \\
\hline
\multicolumn{4}{|c|}{WMT14} \\
\hline
Train.(standard) & 4,500,966 & 113,548,249 & 107,259,529 \\
Dev. (tst2013)  & 3,000 & 64,807 & 63,412 \\
Test (tst2014)  & 3,003 & 67,617 & 63,078 \\
\hline
\multicolumn{4}{|c|}{WMT17} \\
\hline
Train.(standard)    & 4,590,101 & 118,768,285 & 112,009,072 \\
Train.(back trans.) & 10,000,000 & 190,611,668 & 149,198,444 \\
Dev. (tst2016)      & 2,999 & 64,513 & 62,362 \\
Test (tst2017)      & 3,004 & 64,776  & 60,963\\
\hline
\end{tabular}
\end{center}
\caption{\label{tab:dataset} WMT English-to-German corpora}
}
\end{table}

\begin{table}
{\small
\begin{center}
\begin{tabular}{|l|l|}
\hline
\bf Hyperparameter & \bf Value \\
\hline
Embedding Size  & 512 \\
Hidden State Size  & 512 \\
Encoder/Decoder Depth & 2\\
Encoder               & Bidirectional \\
RNN Type              & LSTM \\
Dropout               & 0.2 \\
Label Smooth.          & 0.1 \\
Optimizer             & SGD \\
Learning Rate         & 1.0 \\
Learning Rate Decay   & 0.7 \\
Beam Search Size      & 10 \\
Length Penalty        & 0.6\\
\hline
\end{tabular}
\end{center}
}
\caption{\label{tab:hparam}Hyperparameter Settings}
\end{table}

\subsection{Comparison on Training Speed}
Four baseline toolkits and CytonMT train models using the settings of hyperparameters in Table~\ref{tab:hparam}. The number of layers and the size of embeddings and hidden states varies, as large networks are often used in real-world applications to achieve higher accuracy at the cost of more running time.

Table~\ref{exp:tab:speed} presents the training speed of different toolkits measured in source tokens per second. The results show that the training speed of CytonMT is much higher than the baselines. OpenNMT is the fastest baseline, while CytonMT achieves a speed up versus it by 64.5\% to 110.8\%. Moreover, CytonMT shows a consistent tendency to speed up more on larger networks.  

\begin{table}
{\small
\begin{center}
\begin{tabular}{|l|rrrr|}
\hline
Embed./State Size   & 512 & 512 & 1024 & 1024 \\ 
Enc./ Dec. Layers &  2  & 4   & 2     & 4\\
\hline
Nematus         &  1875 & 1190 & 952  & 604 \\
OpenNMT         &  2872 & 2038 & 1356 & 904 \\
Seq2Seq         &  1618 & 1227 & 854  & 599 \\
Marian          &  2630 & 1832 & 1120 & 688 \\

{\bf CytonMT}         & {\bf 4725} & {\bf 3751} &{\bf 2571} & {\bf 1906} \\
\hline
speedup $\geqslant$        &   64.5\%   & 84.1\%  &  89.6\%  &  110.8\% \\
\hline
\end{tabular}
\end{center}
}
\caption{\label{exp:tab:speed} Training Speed Measured in Source Tokens per Second. }
\end{table}

\subsection{Comparison on Translation Quality}

Table~\ref{tab:bleu} compares the BLEU of CytonMT with the reported results from the systems of the same architecture (attention-based RNN encoder-decoder). BLEU is calculated on cased, tokenized text to be comparable to previous work~\cite{sutskever2014sequence,luong2015addressing,wu2016google,zhou2016deep}.

The settings of CytonMT on WMT14 follows Table~\ref{tab:hparam}, while the settings on WMT17 adopt a depth of 3 and a hidden state size of 1024 as the training set is three times larger.  The cross entropy of the development set is monitored every $\frac{1}{12}$ epoch on WMT14 and every $\frac{1}{36}$ epoch on WMT17, approximately 400K sentence pairs. If the entropy has not decreased by $max(0.01 \times learning\_rate, 0.001)$ in 12 times, learning rate decays by 0.7 and the training restarts from the previous best model. The whole training procedure terminates when no improvement is made during two neighboring decays of learning rate.  The actual training took 28 epochs on WMT14 and 12 epochs on WMT17.

%


Table~\ref{tab:bleu} shows that CytonMT achieves the competitive BLEU points on both benchmarks. On WMT14, it is only outperformed by Google's production system~\cite{wu2016google}, which is very much larger in scale and much more demanding on hardware. On WMT17, it achieves the same level of performance with Marian, which is high among the entries of WMT17 for a single system.   Note that the start-of-the-art scores on these benchmarks have been recently pushed forward by novel network architectures such as \citet{gehring2017convs2s}, \citet{vaswani2017attention} and \citet{shazeer2017outrageously}

\begin{table}
{\small
\begin{center}
\begin{tabular}{|l|c|l|}
\hline
System &  Open Src. &  BLEU \\
\hline
\multicolumn{3}{|c|}{WMT14} \\
\hline
Nematus(Klein,2017)         & $\surd$  & 18.25 \\
OpenNMT(Klein,2017)         & $\surd$ & 19.34 \\
RNNsearch-LV(Jean,2015)    & $\surd$ & 19.4 \\
Deep-Att(Zhou,2016)        &         & 20.6 \\
Luong-NMT(Luong,2015)     & $\surd$ & 20.9 \\
BPE-Char(Chung,2016)        & $\surd$ & 21.5 \\
Seq2seq(Britz, 2017)       & $\surd$ & 22.19 \\
{\bf CytonMT}              & {\bf $\surd$} & {\bf 22.67}   \\
GNMT   (Wu, 2015)                  &  & 24.61 \\
\hline
\multicolumn{3}{|c|}{WMT17} \\
\hline
Nematus(Sennrich,2017) & $\surd$ & 27.5 \\  
{\bf CytonMT}          & {\bf $\surd$ } & {\bf 27.63 } \\
Marian(Junczys,2018)  & $\surd$ & 27.7 \\
\hline
\end{tabular}
\end{center}
}
\caption{\label{tab:bleu} Comparing BLEU with Public Records. }
\end{table}

\section{Conclusion}

This paper introduces CytonMT -- an open-source NMT toolkit -- built from scratch only using C++ and NVIDIA's GPU-accelerated libraries. CytonMT speeds up training by more than 64.5\%, and achieves competitive BLEU points on WMT14 and WMT17 corpora.  The source code of CytonMT is simple because of CytonLib -- an open-source general purpose neural network library -- contained in the toolkit.  Therefore, CytonMT is an attractive alternative for the research community. We open-source this toolkit in hopes of benefiting the community and promoting the field. We look forward to hearing feedback from the community.

The future work of CytonMT will be continued in two directions. One direction is to further optimize the code for GPUs, such supporting multi-GPU. The problem we used to have is that GPUs proceed very fast in the last few years. For example, the microarchitectures of NVIDIA GPUs evolve twice during the development of CytonMT, from Maxwell to Pascale, and then to Volta. Therefore, we have not explored cutting-edge GPU techniques as the coding effort may be outdated quickly. Multi-GPU machines are common now, so we plan to support them.

The other direction is to support latest NMT architectures such ConvS2S~\cite{gehring2017convs2s} and Transformer~\cite{vaswani2017attention}. In these architectures, recurrent structures are replaced by convolution or attention structures. Their high performance indicates that the new structures suit the translation task better, so we also plan to support them in the future.

\bibliography{ref}
\bibliographystyle{acl_natbib_nourl}

\end{document}